\documentclass[acmsmall]{acmart}

\AtBeginDocument{%
  }

\setcopyright{acmcopyright}
\copyrightyear{2018}
\acmYear{2018}
\acmDOI{XXXXXXX.XXXXXXX}

\acmJournal{JACM}
\acmVolume{37}
\acmNumber{4}
\acmArticle{111}
\acmMonth{8}

\usepackage{amsmath,amssymb,amsfonts}
\usepackage{algorithmic}
\usepackage[linesnumbered,ruled]{algorithm2e}
\usepackage{amsthm}
\usepackage{graphicx}
\usepackage{textcomp}
\usepackage{paralist}
\usepackage{subcaption}
\usepackage{hyperref}
\usepackage{multirow}
\usepackage{makecell}
\usepackage{url}
\usepackage{setspace}
\usepackage{tabu}
\usepackage{booktabs} 
\usepackage{romannum}
\usepackage{flushend}
\usepackage{enumitem}
\usepackage{arydshln}
\usepackage[normalem]{ulem}
\usepackage{threeparttable}
\usepackage{ulem}
\usepackage{url}
\usepackage{xcolor}
\definecolor{revised}{RGB}{0,0,255}
\usepackage{soul}
\usepackage{ragged2e}
\usepackage{tabularx}
\usepackage[linewidth=1pt]{mdframed}
\usepackage{cancel}

\begin{document}

\title{Personality-affected Emotion Generation in Dialog Systems}

\author{Zhiyuan Wen}
\email{cszwen@comp.polyu.edu.hk}
\orcid{0000-0001-5644-5023}
\affiliation{%
  \institution{The Hong Kong Polytechnic University}
  \city{Hong Kong}
  \country{China}
}

\author{Jiannong Cao}
\email{csjcao@comp.polyu.edu.hk}
\affiliation{%
  \institution{The Hong Kong Polytechnic University}
  \city{Hong Kong}
  \country{China}
}

\author{Jiaxing Shen}
\authornote{Corresponding author}
\email{jiaxingshen@LN.edu.hk}
\affiliation{%
  \institution{Lingnan University}
  \city{Hong Kong}
  \country{China}
}

\author{Ruosong Yang}
\email{csryang@comp.polyu.edu.hk}
\affiliation{%
  \institution{The Hong Kong Polytechnic University}
  \city{Hong Kong}
  \country{China}
}

\author{Shuaiqi Liu}
\email{cssqliu@comp.polyu.edu.hk}
\affiliation{%
  \institution{The Hong Kong Polytechnic University}
  \city{Hong Kong}
  \country{China}
}

\author{Maosong Sun}
\email{sms@tsinghua.edu.cn}
\affiliation{%
  \institution{Tsinghua University}
  \city{Beijing}
  \country{China}
}

\renewcommand{\shortauthors}{Wen et al.}

\begin{abstract}
Generating appropriate emotions for responses is essential for dialog systems to provide human-like interaction in various application scenarios. Most previous dialog systems tried to achieve this goal by learning empathetic manners from anonymous conversational data.
However, emotional responses generated by those methods may be inconsistent, which will decrease user engagement and service quality. 
Psychological findings suggest that the emotional expressions of humans are rooted in personality traits. 
Therefore, we propose a new task, Personality-affected Emotion Generation, to generate emotion based on the personality given to the dialog system and further investigate a solution through the personality-affected mood transition.
Specifically, we first construct a daily dialog dataset, Personality EmotionLines Dataset (\textbf{PELD}), with emotion and personality annotations. Subsequently, we analyze the challenges in this task, \textit{i.e.},  (1) heterogeneously integrating personality and emotional factors and (2) extracting multi-granularity emotional information in the dialog context. 
Finally, we propose to model the personality as the transition weight by simulating the mood transition process in the dialog system and solve the challenges above.
We conduct extensive experiments on PELD for evaluation. Results suggest that by adopting our method, the emotion generation performance is improved by \textbf{13\%} in macro-F1 and \textbf{5\%} in weighted-F1 from the BERT-base model.
\end{abstract}



\maketitle

\section{Introduction}


Emotional intelligence can be considered a mental ability to reason validly with emotional information and the action of emotions to enhance thought \cite{mayer20042004}. 
Empowering the dialog system with emotional intelligence can enhance plenty of applications like self-disclosure in AI mental therapy, smart customer services, and virtual human in the Metaverse. To achieve this goal, it is necessary to enable the machine to understand users' emotions, generate appropriate emotions, and express them coherently in conversations. 

\begin{figure}[t]
\centering
\includegraphics[scale=0.45]{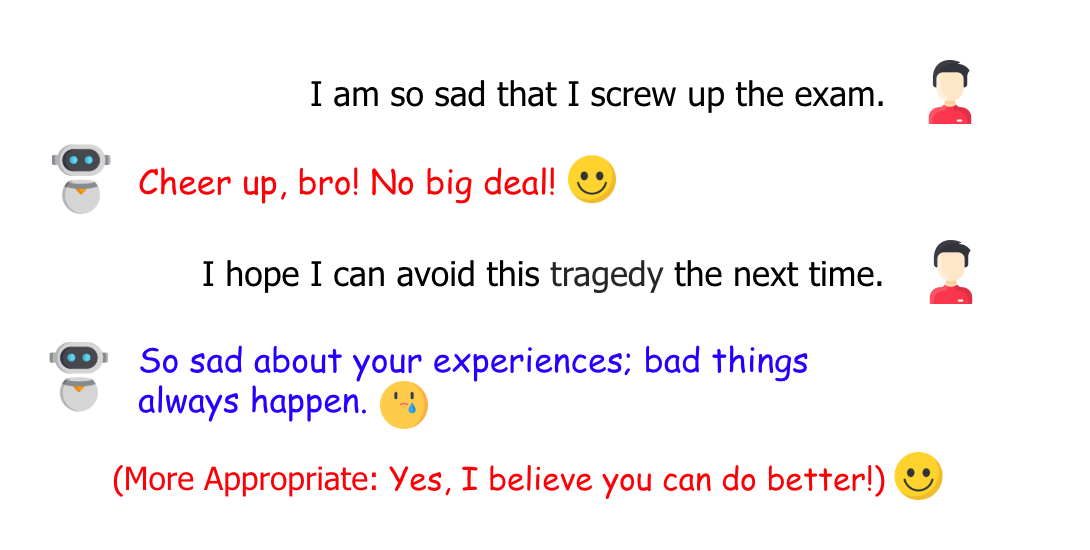} 
\caption{A dialog snippet from a real conversation with a chatbot. The chatbot first optimistically encourages the user with a \textit{Joy} emotion. But after the encouragement, the chatbot pessimistically shows empathy to the user in a \textit{Sadness} emotion even the user expressed a positive attitude. A more appropriate emotional response here should still be in \textit{Joy} so that consistent with the previous response.}
\label{bad_example}
\end{figure}

Despite existing works are relatively mature in recognizing the emotions in dialog context \cite{majumder2019dialoguernn,ghosal2019dialoguegcn,shen2020dual} and rendering specific emotions in responses \cite{zhou2018emotional,colombo2019affect}, the performance of generating appropriate emotions is still limited.
Following the problem settings in empathetic responding \cite{rashkin2018towards}, most existing studies \cite{lin-etal-2019-moel, zandie2020emptransfo, zhong2020towards, zheng2021comae} infer appropriate emotions for the response without modeling the personality of speakers.
Consequently, emotions in responses might be inconsistent with the preceding context. As a result, users may feel they are still talking to rigid machines and reduce their engagement; an example is shown in Figure \ref{bad_example}.

To fill in the research gap, we propose a new task: Personality-affected Emotion Generation, which is to empower the dialog system with personality traits and generate emotion for the response so that the emotions will always be consistent with the given personality trait. Affective instability is a core feature of personality disorder \cite{trull2008affective}, which inspires us that a stable personality trait for the dialog system helps solve the inconsistency in emotional responses. Besides, it is shown that the personality, \textit{i.e.}, the big-five personality model \cite{costa1992normal} can be represented as temperament in the \textbf{V}alence-\textbf{A}rousal-\textbf{D}ominance (VAD) space for emotions \cite{mehrabian1996analysis, mehrabian1996pleasure}.\footnote{It is Pleasure-Arousability-Dominance (PAD) in the original paper, PAD and VAD share the same meaning in the context of text understanding, we will use VAD for consistency henceforth.} These findings suggest that different personalities make different impacts on emotional expressions. Moreover, similar to the one-to-many nature \cite{zhao2017learning} of the dialogue, multiple emotions can be appropriate for response in a similar conversation context, but only one can be selected for the response each time in the dialog system. Accordingly, personality provides an additional reasoning condition to narrow down the searching space so that simplifies the emotion generation.

However, the proposed task entails the following two challenges when applied to reality. 
The first one is heterogeneously integrating personality and emotional factors. Personality plays an important role in inferring appropriate emotions as discussed above. According to the trait theory, personality is usually represented as dimensions or spectrums with extremes at both ends. However, the aspects described by the dimensions are defined in psychological analysis, which is different from the ways we describe the emotional factors in conversations (\textit{e.g.,} discrete emotion labels, or dimensional affective vectors). So, how to integrate the personality with the emotional factors to facilitate our task is difficult.
The second is the extracting multi-granularity emotional information (\textit{i.e.}, emotional factors) from the dialog context. Affective information from conversational data facilitates the emotion generation process. Nevertheless, it can be captured in multiple aspects from conversational data, \textit{e.g.}, utterance-level semantic content, manually annotated emotion labels, and token-level emotional embeddings. Combining affective information in different granularities and different aspects is challenging.

Inspired by personality studies and mood state analysis in human-computer interaction (HCI) \cite{mehrabian1996analysis, itoh2009mood, han2012robotic}, we propose to simulate the mood transition process affected by the given personality traits for emotion generation to solve the above challenges.
In detail, we first project the mood states and emotions of the dialog system as regions and discrete points in the VAD space. 
Then, the transition among the mood states is modeled as the shifting among the different regions, where the emotion factors within the dialog context are the shifting variation, and the personality is modeled as the weight of shifting. In this way, the influence of personality is described in the VAD emotional space where the emotional factors (\textit{e.g.} mood states and emotions)  are also represented. The mood transition process is modeled based on the psychology and HCI domain knowledge, while the parameters are learned from daily conversation data during training. 
To extract the multi-granularity affective information from the dialog context, we design an attention layer to align the semantic representations from BERT \cite{devlin2018bert} as well as the emotion annotations at the utterance level, and the token-level VAD embeddings in different aspects. 
Finally, we observed that different people may express different emotions even in similar mood states or dialog contexts according to their personalities. So, we generate the response emotion (\textit{i.e.}, a specific point corresponding to discrete emotion labels in the VAD space) within the predicted mood state region also conditioned on the given personality trait.


To facilitate related researches, we construct the \textbf{P}ersonality \textbf{E}motion\textbf{L}ines \textbf{D}ataset (PELD), which includes 6,510 dialogue triples from 711 pieces of daily conversations with emotion labels and annotated personality traits. The emotion labels and personality annotations are collected and re-organized from other researchers \cite{poria2018meld,zahiri2017emotion,jiang2019automatic} analyzing the script of a famous TV series \textit{Friends}\footnote{https://en.wikipedia.org/wiki/Friends}. We further explore the mood transition patterns in different characters on PELD and observe that the personality traits influence more in the mood transitions related to negative emotions. Besides, we analyze the distributions of mood state transitions in PELD and the correlations between personality and mood transitions to facilitate future research.

We conduct extensive experiments on the PELD dataset for evaluation. The results verify that by integrating the personality and the mood transition regression modeling, our method achieves significantly better F-scores than the BERT-base model in Emotion Generation, especially for minority emotions. Besides, we also conduct an extensive ablation study to empirically validate the effectiveness of personality and mood transitions in the Emotion Generation task\footnote{Our code and dataset are released at: github.com/preke/PELD.}.

Our key contributions are summarized as follows:

\begin{itemize}
\item To the best of our knowledge, we are the first to raise the effect of personality on generating appropriate emotion in dialog systems. We propose the task of personality-affected Emotion Generation, identify the challenging issues, and propose a solution by simulating the mood transition process in the dialog system inspired by psychological findings.
\item We construct a dialog script dataset PELD with emotion and personality annotations from several existing corpora. Besides, we analyze the distributions of mood state transitions in PELD and the correlations between personality and mood transitions to facilitate future research.
\item We conduct extensive experiments on PELD to evaluate the effectiveness of our method. The results verify that integrating the personality and the mood transition regression significantly improves the F-scores than base models in Emotion Generation, especially in minority emotions.
\end{itemize}

In the following content, we first review the related studies in Section~\ref{section:Related work}. Section~\ref{section:Preliminaries} describes the background and preliminary knowledge of our work. Section~\ref{section:Methodology} presents the problem definition of personality-affected emotion generation and the methodology we propose based on mood transition prediction. Section~\ref{section:Experiment} introduces the experimental setup. Section~\ref{section:Result} describes the evaluation results and analysis. Finally, we conclude and mention future work in Section~\ref{section:Conclusion}.

\section{Related Works}
\label{section:Related work}
To enhance the understanding of the context behind our study and illustrate the distinctiveness of our work compared with prior research, we conducted a comprehensive review of the current literature on emotional dialog systems and personality influence on emotion expression.

\subsection{Emotional Dialog Systems}
The idea of an emotional dialog system was first proposed in Colby's 1975 publication \cite{colby1975artificial}, where a rule-based chatbot for simulating emotions was introduced. Present research on emotional dialog systems predominantly concentrates on three areas: (1) emotion recognition in conversations, (2) emotional response generation, and (3) empathetic response generation.

\subsubsection{Emotion Recognition in Conversations}
The ERC task aims to identify the emotion of each utterance from several pre-defined emotions given the transcript of a conversation along with speaker information on each constituent utterance \cite{poria2019emotion}. Relevant research is becoming increasingly popular recently, largely because of its potential to extract opinions from the enormous amounts of conversational data accessible on diverse online platforms such as Facebook, YouTube, Reddit, Twitter, and other similar sites\cite{poria2019emotion}. Conversational memory network (CMN), proposed by Hazarika et al. \cite{hazarika2018conversational} for emotion recognition in conversation, utilizes distinct memories for each speaker for speaker-specific context modeling. Later, Hazarika et al. \cite{hazarika2018icon} improved upon this approach with an interactive conversational memory network (ICON), which interconnects these memories to model self and inter-speaker emotional influence. However, neither methods actually exploit the speaker's information of the target utterance for classification. This makes the model blind to speaker-specific nuances. DialogueRNN \cite{majumder2019dialoguernn} aims to solve this issue by considering the speaker's information of the target utterance and, further, modeling self and inter-speaker emotional influence with a hierarchical multi-stage RNN with the attention mechanism. Dialogue Graph Convolutional Network (DialogueGCN) \cite{ghosal2019dialoguegcn}, a graph neural network-based approach is proposed by leveraging self and inter-speaker dependency of the interlocutors to model conversational context for emotion recognition. Through the graph network, DialogueGCN addresses context propagation issues present in the current RNN-based methods.

\subsubsection{Emotional Response Generation}
In dialog systems, emotional response generation is mainly concerns with incorporating emotional information into responses \cite{ma2020survey}. Research related to emotional response generation has gained popularity recently, following Microsoft's introduction of Xiaoice, a chit-chat robot capable of recognizing users' emotional needs, back in 2014 \cite{zhou2020design}. Emotional Chatting Machine \cite{zhou2018emotional} was proposed to exploit the deep learning approach in building a large-scale emotionally aware conversational bot. Emotional embeddings are also trained based on context and then integrated into response generation \cite{shantala2018neural}. Some researchers \cite{colombo2019affect} control the emotional response generation with both categorical emotion representations and continuous word representations in VAD space \cite{mohammad2018obtaining}. Moreover, affectively diverse beam search \cite{asghar2018affective} is proposed for decoding in response generation. Besides, reinforcement learning is also adopted to encourage response generation models to render specified emotions. Li et al. \cite{li2019reinforcement} combined reinforcement learning with emotional editing constraints to generate meaningful and customizable emotional replies. Sun et al. \cite{sun2018emotional} also use an emotion tag to partially reward the model for expressing specified emotion. 

\subsubsection{Empathetic Dialogue Generation}
Despite the existing advancement in emotion response generation, it is impractical always to specify response emotions for dialog systems in real application scenarios. Therefore, empathetic dialogue generation aims to understand the emotion of users and respond to them appropriately. Rashkin et al. \cite{rashkin2018towards} propose a benchmark for empathetic dialogue generation and a dataset (EMPATHETICDIALOGUES) of 25k conversations grounded in emotional situations. Followingly, the Mixture of Empathetic Listeners (MoEL) \cite{lin-etal-2019-moel} was proposed as a novel end-to-end approach for modeling empathy in dialogue systems. Their model first captures the user's emotions and outputs an emotion distribution. Based on this, MoEL will softly combine the output states of the appropriate Listener(s), which are each optimized to react to certain emotions and generate an empathetic response. It is stated that empathetic responses often mimic the emotion of the user to a varying degree, depending on its positivity or negativity and content \cite{majumder2020mime}. So, they use emotion stochastic sampling and emotion mimicry to generate responses that are appropriate and empathetic for positive or negative statements. EmpTransfo \cite{zandie2020emptransfo} utilizes OpenAI-GPT for language generation. The authors also show that the history of emotions and other metadata can improve the quality of generated conversations by the dialog system. To understand expressed empathy in text-based, asynchronous conversations, Sharma et al. \cite{sharma2020computational} develop a multi-task RoBERTa-based \cite{liu2019roberta} bi-encoder model for identifying empathy and extracting rationales underlying its predictions. Zheng et al. \cite{zheng2021comae} state that empathy expression is multi-dimensional and is influenced by various factors. Therefore they propose CoMAE modeling communication mechanism, dialog act, and emotion together for empathetic response generation. To simulate the emotional interaction among humans, Wei et al. \cite{wei2019emotion} design an emotion selector to learn the proper emotion for responses from massive dialogue pairs. But the emotional expression is subjective; for the same post, different users may have different emotions in their responses. So, the pattern learned only from online dialogues ignores the user information and turns out to be impractical.

\subsection{Personality Effects on Emotions}
Besides existing studies in emotional dialog systems, some researchers in related disciplines also consider projecting the effect of the personality of emotions in designing conversation agents. Research about personality effects on emotions is mainly distributed in cognitive science and Human-Computer Interaction (HCI). Due to the limited dialog corpus with personality annotations, related works in text-based conversation are few. However, some researchers also investigate how factual persona information influences emotion expressions in conversation.

\subsubsection{Personality Effects on Emotions in Cognitive Science}
Personality affects the way of communication in various manners including both linguistic style \cite{mairesse2011controlling} and acoustic traits \cite{polzehl2016personality}. As it feels more natural to interact with a machine that has its own personality, implanting personality into dialogue agents would possibly increase the social attachment \cite{lee2006can}. While emotion is a complex psychological experience of an individual’s state of mind as interacting with people or environmental influences \cite{han2012robotic}. The \textbf{P}leasure-\textbf{A}rousal-\textbf{D}ominance (PAD) \cite{mehrabian1996pleasure} or \textbf{V}alence-\textbf{A}rousal-\textbf{D}ominance (VAD) emotion temperament model shows three nearly orthogonal dimensions providing a comprehensive description of emotional states. Based on this, several psychologists studied the relationship between human emotional factors and personality factors. However, most of them are rule-based models \cite {johns2001emotions} and probabilistic models \cite{andre1999integrating}. Mehrabian et al. \cite{mehrabian1996analysis} utilized the five factors of personality  \cite{costa1992normal} to represent the VAD temperament model through linear regression analysis. 

\subsubsection{Personality Effects on Emotions in HCI}
To integrate the analysis above into Artificial Intelligence, some researchers in HCI borrow the ideas above and design facial emotional expressions for humanoid robots. An emotion generation model \cite{itoh2009mood} was proposed that represents a robot's internal state. This model can assess the robot's individuality through mood transitions. The mood transition process of robots was also introduced in this work. Later on, a method of the smooth transition of emotional states for a robotic face was developed by Han et al. \cite{han2010design}. The proposed system can generate emotional behaviors of a robot in a more natural manner based on the 2-D emotional model. Subsequently, they continued to employ five factors of personality in a 2D (pleasure-arousal) scaling model to represent a robotic non-verbal emotional model \cite{han2012robotic}. Based on previous works, Masuyama et al. \cite{masuyama2015robotic} studied the three stages (core affects, emotion, and mood state) robotic emotional model with the OCEAN model as the personality factors based on the 2D (Pleasant and Arousal) scaling model. Then, a personality-affected robotic emotional model and the emotion-affected associative memory model for the robot \cite{masuyama2018personality} are introduced for robots expressing emotions. The robots in this work provide non-verbal emotional interaction with users where the pre-defined personalities of robots affect their propensity for simulated mood transitions. 

\subsubsection{Effects of Persona Information on Emotions}
While in the research of text-based dialog systems, early work in \cite{ball2000emotion} utilizes models of emotions and personality encoded as Bayesian networks to generate empathetic behaviors or speech responses to users in conversation. Though the VAD space is adopted to model emotions in some research \cite{colombo2019affect,asghar2018affective,wen2021automatically}, integrating the psychological personality into the dialog system is rare due to the data shortage for personality analysis in conversation. Consequently, the psychological personality influence on emotion in dialogues is still an open problem. It is worth mentioning that, Zhong et al. propose a new task for persona-based empathetic conversations and present the first empirical study on the impact of persona on empathetic responding \cite{zhong2020towards}. The authors use the description of persona as the additional input to generate empathetic responses based on the statement that persona has been shown to be highly correlated to personality, which in turn influences empathy. Although the impact of persona on generating appropriate emotions is validated in the results, as a more straightforward factor to affect emotion generation, the influence of personality in emotional dialog systems has not been studied before.

\section{Preliminaries}
\label{section:Preliminaries}

In this section, we introduce some concepts from psychology and affective computing for a better explanation of our problem and method.

\subsection{Personality, Mood-state, and Emotion}
Emotion (short-term), mood-state (mid-term), and personality (long-term) thus represent three levels of the affective module that interact with each other \cite{kessler2008simplex}. Specifically, emotion is the term used in psychology to describe short-term variations in internal mental state \cite{ball2000emotion}. Beyond discrete emotions, which are typically short-term, mood states are a powerful way to model emotional shifts and explain affective influences over longer periods of time \cite{kessler2008simplex}. Personality characterizes the long-term patterns of thought, emotion, and behavior associated with an individual \cite{ball2000emotion,wen2023desprompt}. 

\begin{table}[h]
    \centering
    \caption{Numeric vectors of Basic Emotions in the VAD Space \cite{russell1977evidence}.}
    \linespread{1.2}
    \small
    \begin{tabular}{c|c}
        \toprule  
        \textbf{Basic Emotions} & \textbf{(Valence, Arousal, Dominance)}\\
        \hline  
		Anger & (-0.51, 0.59, 0.25) \\		
		Disgust & (-0.60, 0.35, 0.11) \\
		Fear & (-0.62, 0.82, -0.43) \\
		Joy & (0.81, 0.51, 0.46)\\
		Neutral & (0.00, 0.00, 0.00) \\
		Sadness & (-0.63, -0.27, -0.33) \\
		Surprise & (0.40, 0.67, -0.13) \\
		\bottomrule
    \end{tabular}
    \label{Emo_mapping}
\end{table}

\subsection{Basic Emotions in the VAD Space}
There are two fundamental approaches to model emotions in psychological and affective computing literature: categorical and dimensional models. Categorical emotion models assume that all humans have a given discrete set of basic emotions (\textit{i.e.}, \textit{Anger}, \textit{Disgust}, \textit{Fear}, \textit{Joy}, \textit{Sadness}, and \textit{Surprise} \cite{ekman1994nature}). These models are easily adopted for classification tasks in emotion analysis problems \cite{ghosal2019dialoguegcn,wang2020contextualized}. However, different researchers propose various standards of categorization \cite{russell1991culture}, making it difficult to unify emotion analysis research. Dimensional approaches, on the other hand, often refer to Russell and Mehrabian's Valence-Arousal-Dominance (VAD) model  \cite{mehrabian1996pleasure}. According to this model, emotional states can be described relative to three fundamental emotional dimensions: Valence (the degree of pleasure or displeasure of an emotion), Arousal (level of mental activity, ranging from low engagement to ecstasy), and Dominance (extent of control felt in a given situation). Accordingly, emotions are characterized by three dimensions, each of which spans an interval of real-valued numbers indicating the strength and orientation of each dimension  \cite{buechel2016emotion}. 

To better utilize the strength in different approaches, discrete emotions are projected into the dimensional space as shown in Table \ref{Emo_mapping}\footnote{\textit{Fear} and \textit{Joy} correspond to \textit{Terrified} and \textit{Happy} in the original reference.}. Consequently, a wide range of existing emotion analysis models can be embedded into the VAD space \cite{li2017inferring,li2017affective,bauerhenne2020emotional}. Besides, emotions in the VAD space can have multiplex comparisons. For example, all \textit{Anger}, \textit{Disgust} and \textit{Fear} are negative emotions according to the valence dimension, but the intensity of \textit{Anger} is higher than \textit{Disgust} and \textit{Fear} referring to the arousal dimension.

\begin{table}[b]
    \centering
    \caption{The OCEAN personality traits and description \cite{costa1992normal}.}
    \small
    \begin{tabular}{ll}
        \toprule  
        \textbf{Factor} & \textbf{Description}\\
        \midrule  
		Openness & Openminded, imaginative, and sensitive. \\		
		Conscientiousness & Scrupulous, well-organized.\\
		Extraversion & The tendency to experience positive emotions. \\
		Agreeableness & Trusting, sympathetic, and cooperative. \\
		Neuroticism	&  The tendency to experience psychological distress. \\
		\bottomrule
    \end{tabular}
    
    \label{ocean traits}
\end{table}

\subsection{Mood States in the VAD Space}
\begin{figure}[t]
\centering
\includegraphics[scale=0.32]{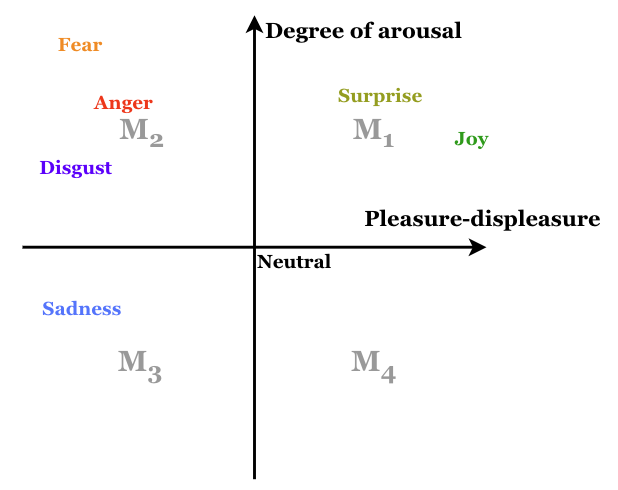} 
\caption{Mood domains and emotions in the VAD space. We only illustrate the plane of Valence (Pleasure-displeasure) and the Arousal (Degree of arousal) axis here because the mood domains are split only by these two dimensions \cite{itoh2009mood}.  $M_1$, $M_2$, $M_3$, and $M_4$ are different quadrants representing different mood states.}
\label{mood_vad}
\end{figure}

Mood states provide the historical context of recent emotions \cite{marinier2007computational}, and they are not usually characterized by their direction at a person or event. While someone might show an emotion \textit{Anger} toward a specific object (\textit{e.g.}, a colleague), as the specific emotion dissipates, they may feel a generally bad mood \cite{wagner2017psychological}. Mehrabian et al. \cite{mehrabian1996pleasure} give specific names to the resulting different octants in VAD space and describes the diagonally opposite octants as Exuberant/Bored, Dependent/Disdainful, Relaxed/Anxious, Docile/Hostile. Thus mood-states are not points but octants of the VAD space. More practically, moods can also be represented in four categories by dividing the VAD spaces into four domains through the axis of pleasure-displeasure and degree of arousal, as shown in Figure \ref{mood_vad}. Emotions from $M_1$ (\textit{e.g., Joy}) are with positive polarity and high intensity. While \textit{Sadness} from $M_3$ is with negative polarity and low intensity, for instance.

\subsection{Personalities in the VAD Space}
The big five personality traits (OCEAN, shown in Table \ref{ocean traits}) are widely used for psychological analysis. Higher scores for openness are correlated with philosophical and free thought, as well as an interest in the arts, music, and cinema \cite{schwartz2013personality,friedman2014personality}. Those who score low here may be more practical, realistic, or close-minded \cite{costa1992normal}. Individuals with high conscientiousness tend to be well-organized, which may be expressed through discussions of work or school-related responsibilities \cite{yarkoni2010personality, friedman2014personality}. Those who score lower on this dimension may appear impulsive or unreliable. Those with high extraversion are likely to talk about friends and are willing to have interpersonal interaction. However, low extraversion people may be more independent and may focus more on solo activities \cite{costa1992normal, park2015automatic}. Agreeableness is associated with being friendly and good-natured, while those who score low may be selfish or rude. Swearing is highly correlated with low agreeableness \cite{yarkoni2010personality,schwartz2013personality}. High neuroticism is strongly linked to anxiety and depression, while low neuroticism is linked to emotional stability. This dimension may be expressed through feelings such as fear, sadness, or frustration \cite{costa1992normal,friedman2014personality}.

Positioning a personality within a VAD space could have been a rather difficult task since there is no mathematically-correct way to make the conversion. Luckily, this transformation can be based on empirical data. \cite{mehrabian1996analysis} proposed a temperament model shown in Equation \ref{personality} derived through linear regression to show the VAD scales of personality traits, where $O, C, E, A, N$ are the strength of the big-five personality traits.

\begin{equation}
	\begin{aligned}
	P_V &= 0.21E + 0.59A + 0.19N \\
	P_A &= 0.15O + 0.30A - 0.57N \\
	P_D &= 0.25O + 0.17C + 0.60E - 0.32A \\
	\end{aligned}
	\label{personality}
\end{equation}

\noindent
It is seen that VAD scores can be estimated with moderate accuracy when only big-five scores are available, and investigators also wish to analyze their data in relation to the VAD temperament model.

\section{Personality-affected Emotion Generation}
\label{section:Methodology}
\subsection{Problem Statement}

We define Personality-affected Emotion Generation as enabling the dialog system to generate appropriate emotions for response facilitated by the given personality trait. 

Formally, we study a dyadic emotional conversation between the user and the dialog system, which contains the dialog context $C=\{(U_1, E_1), (U_2, E_2), ..., (U_{n-1}, E_{n-1})\}$ including all the preceding $n-1$ utterances, where  $E_i$ is the emotion label for each utterance $U_i$. The personality trait $P$ of the dialog system is also given. Our objective is to let the dialog system generate an appropriate emotion $E_n$ for the upcoming response $U_n$ to the user. We evaluate the appropriateness of the generated $E_n$ by comparing it with the ground truth emotion $E_n\prime$ in the dialog corpus. $E_n\prime$ is generated by the speaker annotated with the same personality $P$ in the given context.

\begin{figure*}[t]
\centering
\includegraphics[width=380pt, height=200pt]{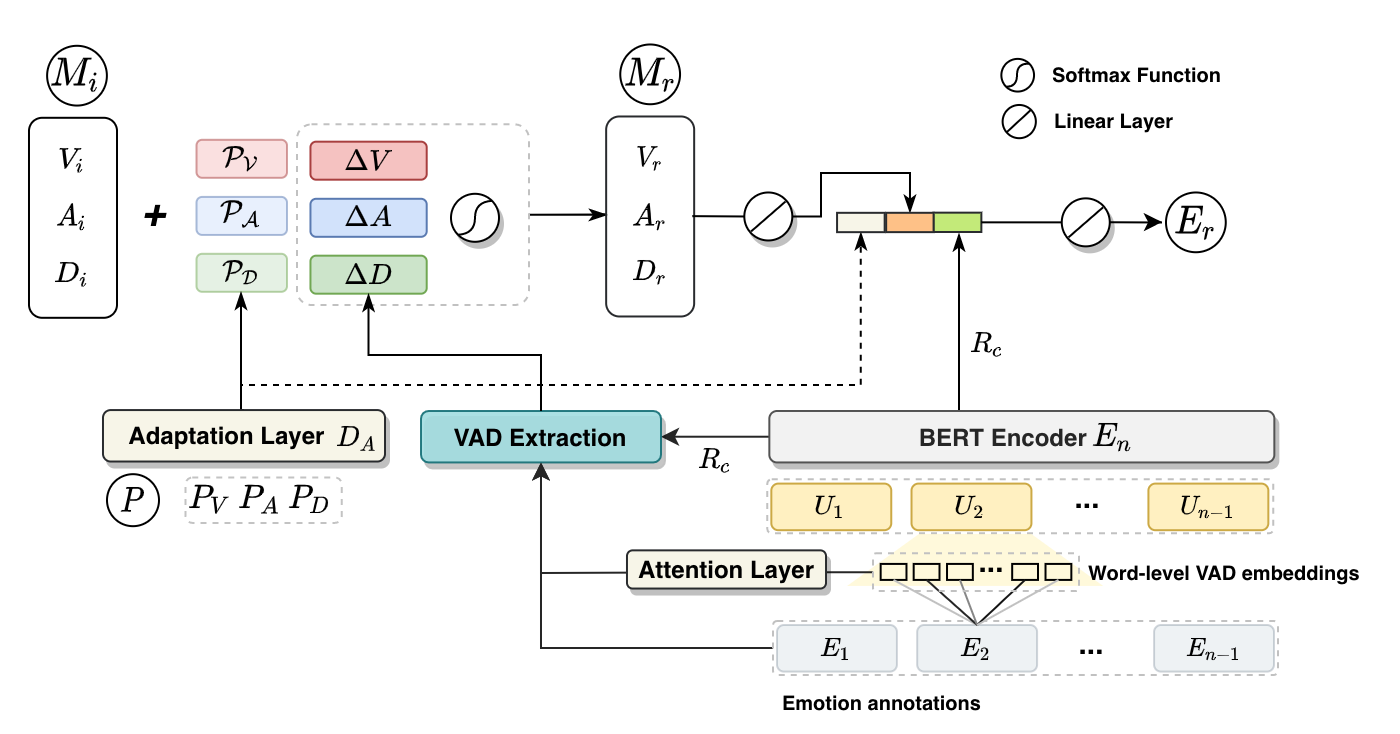} 
\caption{The Model Illustration. The upper-left part is the mood transition regression, where the initial mood state $M_i$ changes to the $M_r$ in the upcoming response. The personality trait $P$ is transformed to the transition weights $\mathcal{P_V, P_A, P_D}$, and the extracted affective information from the dialog context is changed to the transition variables $\Delta V, \Delta A, \Delta D$. Then, the upper-right part is the emotion generation module, where the response $E_r$ is generated through integrating $M_r$, $P$, and the dialog context $R_c$. The lower part illustrates how we extract affective information by the Attention Layer aligning the emotion annotations, word-level VAD embeddings, and the semantic representations from the input dialog context.}
\label{model}
\end{figure*}

Based on the problem definition and the preliminaries above, we design the Personality-affected Mood Transition model as illustrated in Figure \ref{model}. Our model includes two modules: Mood Transition Regression and Emotion Generation. We will introduce both modules in the following subsections.

\subsection{Mood Transition Regression}
\subsubsection{Personality Effect on Mood Transition}

Mood dynamically changes with the accumulation of past emotions. Changing the mood transition of robots can produce various characterizations in them as well as individuality that enables the robots to demonstrate a wide range of behaviors \cite{itoh2009mood}. Moreover, the transition of mood depends not only on the user’s emotions but also on the mood and personality of the dialog system\cite{han2012robotic}. Therefore, we first model the personality of the dialog system as the weighting parameters of the mood state transition.

In our model, the mood states are represented by $M_1, M_2, M_3, M_4$, and $Neutral$ referring to Figure \ref{mood_vad}, where the $Neutral$ especially represent the coordinate origin of the VAD space\footnote{We generally regards $M_1, M_2, M_3, M_4$, and $Neutral$ are different regions (octants) in the VAD space}.  We use mood vectors in Table \ref{Mood regions} to represent the mood states in our model. Therefore, the mood transition is the shifting among different octants in the VAD space \cite{walter2013transsituational}.

Besides, the personality of the dialog system is specified as a 5-dimensional vector $P = [O, C, E, A, N]$ in our method representing the strength in Openness, Conscientiousness, Extraversion, Agreeableness, and Neuroticism, respectively. 
The temperament of personality in the VAD space (shown in Equation \ref{personality}) is widely used as the weighting parameter for mood transition of robots \cite{han2012robotic,masuyama2018personality,bauerhenne2020emotional}, which is also adopted by us as the preliminary personality traits of the dialog system.

\begin{table}[h]
    \centering
    \caption{Mood VAD vectors representing different mood states}
    \linespread{1.2}
    \small
    \begin{tabular}{c|c}
        \toprule  
        \textbf{Mood States} & \textbf{(Valence, Arousal, Dominance)}\\
        \hline  
		$M_1$ & (1.0, 1.0, 0.0) \\		
		$M_2$ & (-1.0, 1.0, 0.0) \\		
		$M_3$ & (-1.0, -1.0, 0.0) \\		
		$M_4$ & (1.0, -1.0, 0.0) \\		
		Neutral & (0.00, 0.00, 0.00) \\
		\bottomrule
    \end{tabular}
    \label{Mood regions}
\end{table}

In detail, we first obtain $P_V, P_A, P_D$ referring to Equation \ref{personality} as the preliminary affecting parameters. However, these numeric coefficients are summarized from the analysis of questionnaire results from 72 participants \cite{mehrabian1996analysis}, which are not suitable for directly adopted in the model design.
Our intent is to learn how personality affects the mood transition process in conversation from the dialog corpus. Therefore, we choose $P_V, P_A, P_D$ as prior parameters and learn suitable coefficients $\mathcal{P_V, P_A, P_D}$ from conversational data through an adaptation dense layer $D_A$.  

\subsubsection{Affective Information Extraction}
In human-robot interaction, users' emotional expressions can be treated as trigger inputs to drive the robotic mood transition \cite{han2012robotic}. Inspired by this idea, we extract the affective information from the utterances in the dialog context to facilitate the mood transition process in our model. 

Extracting the affective information from utterances has been studied by previous researchers. Buechel et al.\cite{buechel2017emobank} annotate a corpus of over 10,000 English sentences employing the dimensional VAD model of emotion. Park et al. \cite{park2019dimensional} predict VAD vectors with the supervision of categorical emotion annotations by minimizing the EMD (Earth Mover’s Distance) loss between the predicted VAD score distribution and the categorical emotion distributions. However, both methods overlook the affective information in the semantic content. So, we design an affective extraction module to integrate the semantic content, token-level VAD embeddings, and emotion annotations for each utterance together.

Specifically, at the utterance level, we first fine-tune the pre-trained BERT\footnote{Here we adopt the pre-trained BERT-base model from Huggingface (https://huggingface.co/).} \cite{devlin2018bert} encoder $E_n$, a famous pre-trained language model whose performance is widely validated in many natural language understanding tasks, on daily conversation corpus by concatenating all the utterances in $C$ with the [SEP] to form the input and obtain the semantic representation $R_c$ from the dialog content. 
\begin{equation}
	\begin{aligned}
	R_c &= E_n(U_1 [SEP] U_2,...,U_{n-1}[SEP]) 
	\end{aligned}
	\label{r_c}
\end{equation}
Then, at the token-level, for each utterance $U_i$ in $C$, we design an attention layer using the VAD embeddings $[V_i, A_i, D_i]$ of the emotion annotation $E_i$ to attend the VAD embeddings of all the tokens $[t_1, t_2, ..., t_m]$ in $U_i$. 
The VAD embeddings for the emotion annotations and each word in the utterance are obtained by the manual VAD annotations for English words \cite{mohammad2018obtaining}. The attention weights $W_j$ are calculated by the softmax function over the inner product between the VAD embeddings of $E_i$ and $t_j$: 
\begin{equation}
	\begin{aligned}
	W_j &= \frac{e^{E_i^Tt_j}}{\sum_{j=1}^{m}e^{E_i^Tt_j}}\\\
	A_i &= \sum_{j=1}^{m} W_jt_j
	\end{aligned}
	\label{attention}
\end{equation}

Our principle here is that not all words are affected by emotion. For example, function words are mostly only affected by syntax while topical/factual words are dominated by topics or facts \cite{ma2020survey,wen2020decode}. Therefore, a higher attention weight means the current word is more important in expressing emotion. The attention output $A_i$ combines the token-level (word VAD embeddings) and the utterance-level (utterance emotion annotations) affective information for each sentence. 

Finally, we sum the average among all the $E_i$ for each utterance, and the average among all the output $A_i$ from the attention layer together as the affective information extracted from the dialog context. Besides, the semantic representation $R_c$ is also concatenated with the affective information and is fed into a linear layer to obtain the mood variation $\Delta V, \Delta A, \Delta D$.
\begin{equation}
	\begin{aligned}
	E &= \sum_{i=1}^{n-1} \frac{E_i}{n-1} \\\
	A &= \sum_{i=1}^{n-1} \frac{A_i}{n-1} \\\
	\Delta V, \Delta A, \Delta  D &= Linear((w_EE + w_AA)\oplus R_c)
	\end{aligned}
	\label{delta_vad}
\end{equation}
where $w_E$ and $w_A$ are hyperparameter weights to balance the two aspects.

\subsubsection{Mood State Regression}
We utilize the regression methods to predict the response mood $M_r$, which aims to describe the distance among different mood states in the VAD space. Firstly, we obtain the numeric values of the $M_r$ by Table \ref{Mood regions}. The predicted mood state $M_r\prime: V_r^\prime, A_r^\prime, D_r^\prime$ is calculated by adding the preceding mood state $M_i : V_i, A_i, D_i$ and the variation of the mood weighted by the personality coefficients $\mathcal{P_V, P_A, P_D}$. What is more, we add a softmax function to the mood variation $\Delta V, \Delta A, \Delta D$ to constrain their numeric scale into $[0-1]$ for the transition regression.

\begin{equation}
	\begin{aligned}
	V^\prime_r &= V_i + \mathcal{P_V} * Softmax(\Delta V) \\\
	A^\prime_r &= A_i + \mathcal{P_A} * Softmax(\Delta A) \\\
	D^\prime_r &= D_i + \mathcal{P_D} * Softmax(\Delta D) 
	\end{aligned}
	\label{vad_calc}
\end{equation}

\subsection{Emotion Generation} 
After predicting the mood state of the upcoming response, we generate the appropriate emotion that the dialog system responds to the user. The dialog context acts as a set of parameters that influence a person to speak an utterance while expressing a particular emotion \cite{poria2018meld}. Besides, emotion expressions are also affected by the personality trait.
 
According to our observation, different people may express different emotions in similar mood states or dialog contexts according to their personalities. For example, under a stressful context, individuals who exhibit high levels of Neuroticism will tend to be irritable and express anxiety or anger, while a person who ranks lower on the neurotic level will peacefully exhibit emotions with lower arousability.

Therefore, we concatenate the predicted mood state $M^\prime_r$, the personality trait $P$, and the semantic representation $R_c$ of the dialog context as the input. Here, to project the mood state and the personality trait into the same feature space as $R_c$, we adopt two linear layers $Linear_{m}$ and $Linear_{p}$ respectively on $M^\prime_r$ and $P$.
Then, the $E_r$ is generated from $M^\prime_r$ through a linear classification layer $Linear_{cls}$ conditioned on both $R_c$ and $P$, as described in Formula \ref{emo_gen}. 
\begin{equation}
\begin{aligned}
E_r &= Linear_{cls}(Linear_m(M^\prime_r) \oplus Linear_p(P_n) \oplus R_c) \\
\end{aligned}
\label{emo_gen}
\end{equation}

\subsection{Joint Training}
As the mood transition prediction part is the preceding module of the emotion generation part, we jointly optimize the mood transition prediction part and the emotion generation part during training so that the performance of the latter part could be enhanced.

First, we calculate the mean square error (MSE) as the mood loss $\mathcal{L}_{mood}$ between the regression results in Formula \ref{vad_calc} and the mood state vectors $[V_r, A_r, D_r]$ of actual $M_r$ in Table \ref{Mood regions}.
\begin{equation}
\begin{aligned}
\mathcal{L}_{mood} &= \sqrt{(V_r - {V^\prime_r})^2} + \sqrt{(A_r - {A^\prime_r})^2} +\sqrt{(D_r - {D^\prime_r})^2}
\end{aligned}
\label{mse_loss}
\end{equation}

For the emotion generation, based on our observation of dialog corpus that discrete emotions are unequally distributed in daily conversations (\textit{e.g.}, \textit{Neutral} is the majority among all the emotions, while \textit{Fear, Disgust} are the minority), we use the Focal Loss\cite{lin2017focal} between the generated emotion and the ground truth emotion to address class imbalance during training, as shown in \ref{focal_loss}. 

Focal Loss applies a modulating term to the cross-entropy loss in order to focus learning on hard misclassified examples. It is a dynamically scaled cross-entropy loss, where the scaling factor decays to zero as confidence in the correct class increases. Intuitively, this scaling factor can automatically down-weight the contribution of easy examples during training and rapidly focus the model on hard examples. Formally, the Focal Loss adds a factor $(1-p_t)^\gamma$ to the standard cross-entropy criterion, where the $p_t$ is the probability of generating the correct emotion. Setting $\gamma > 0$ reduces the relative loss for well-classified examples $(p_t > 0.5)$, putting more focus on hard, misclassified examples.
\begin{equation}
\begin{aligned}
\mathcal{L}_{emo} &= FL(p_t) = -\alpha (1-p_t)^\gamma log(p_t)
\end{aligned}
\label{focal_loss}
\end{equation}
The $\alpha$ is set to be $0.5$ and $\gamma$ to be $2$ according to the empirical results in the original paper \cite{lin2017focal}. 

Finally, we minimize the mood loss and emotion loss simultaneously with adaptive weights $w_1, w_2$ learned during training.
\begin{equation}
\begin{aligned}
\mathcal{L} = w_1\mathcal{L}_{mood} + w_2\mathcal{L}_{emo}
\end{aligned}
\label{loss}
\end{equation}

\section{The PELD Dataset}
\label{section:Dataset}
\subsection{Dataset Construction \& Statistics}

\begin{figure}[t]
\centering
\includegraphics[scale=0.45]{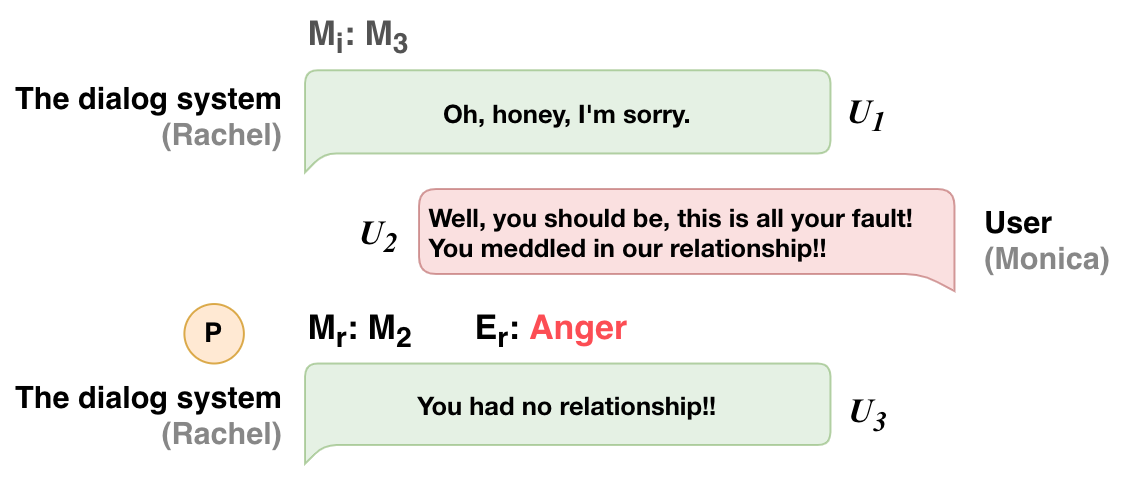} 
\caption{A triple example in PELD. The dyadic conversation between Rachel and Monica (two main roles in \textit{Friends}, $P$ is the personality of Rachel. In this example, the dialog system is set as Rachel and generate the emotion \textit{Anger} in the response $U_3$ to the user set as Monica.}
\label{triple}
\end{figure}

To facilitate related research, we construct the \textbf{P}ersonality \textbf{E}motion\textbf{L}ines \textbf{D}ataset (\textbf{PELD}), an affective dialog dataset with personality traits for speakers and emotion annotations for utterances. In PELD, each sample is represented as a dyadic dialog triple $\{ (U_1, U_2, U_3), M_i, M_r, E_r, P$\}, as shown in Figure \ref{triple}. $M_i$ and $M_r$ are mood states expressed in $U_1$ and $U_3$, $P$ is the personality trait, and $E_r$ is the emotion label to be generated. 

According to the book \textit{Television Dialogue: The sitcom Friends vs natural conversation} \cite{quaglio2009television}, television conversations and natural conversations are basically the same in terms of linguistic features. This classic script is widely analyzed in many dialog researches \cite{li2016persona,li2020transformers,jiang2019automatic}. Therefore, we construct the triples in PELD from the dialog scripts in Friends based on existing studies. 

Specifically, the utterances and their emotion labels are mainly adopted from the dialogues in the MELD \cite{poria2018meld} and the EmoryNLP dataset \cite{zahiri2017emotion}, two famous datasets analyzing emotional expressions in \textit{Friends}.
The VAD vector of each mood state refers to Table \ref{Mood regions}. To keep consistency, each dialog triple in PELD is constructed within the same dialogue in the original datasets.  
The personality traits in our dataset are adopted from the personality annotations in 711 different dialogues \cite{jiang2019automatic}. We only keep the personality traits of the six main roles in \textit{Friends} for confidence, as these annotations are most frequent. Referring to the annotations, a role may exhibit different aspects of its personality in different conversation scenarios. Thus, for each of the main roles, we average their annotated personality traits in all the dialogues by $P = \frac{1}{K}\sum_{i=1}^K{P_i}$ for simplification, where $K$ is the number of annotations. The averaged results are shown in Table \ref{personality_}.

%


\begin{table}[h]
    \centering
    \caption{Personality vectors of \textit{Friends} main roles in PELD.}
    \linespread{1.2}
    \small
    \begin{tabular}{l|l}
        \toprule  
       \textbf{Roles} & \textbf{Personality Traits (O,C,E,A,N)}\\
        \hline
		Chandler & [0.648, 0.375, 0.386, 0.58, 0.477]\\
		Joey & [0.574, 0.614, 0.297, 0.545, 0.455]\\
		Monica & [0.713, 0.457, 0.457, 0.66, 0.511]\\
		Phoebe & [0.6, 0.48, 0.31, 0.46, 0.56]\\
		Rachel & [0.635, 0.354, 0.521, 0.552, 0.469]\\
		Ross & [0.722, 0.489, 0.6, 0.533, 0.356]\\
	    \hline
	    \textbf{Std} & [0.059, 0.093, 0.120, 0.065, 0.068] \\
		\bottomrule
    \end{tabular}
    \label{personality_}
\end{table}

We also calculate the standard deviation of the personality traits on each dimension. It suggests that Extroversion and Conscientiousness share the largest distinction among the five traits in the six main roles. These two traits, especially Extroversion, play important roles in mood state variation and emotion generation according to the previous discussion. Therefore, it also reflects the availability of PELD in the aspect of personality.

We split the PELD into \textbf{Train}, \textbf{Valid}, and \textbf{Test} set with portion around 8:1:1. There are 6,527 triples in PELD. The total number of unique utterances in PELD (10,468) is less than the sum of the original MELD (13,708) and the EmoryNLP (9,489). The reason is there are lots of overlap between these two datasets. Besides, not all dialogues include the six main roles and are suitable for constructing triples. 
The overall statistics of the dataset are shown in Table \ref{PELD_dataset}\footnote{According to the annotations in PELD, there is no utterance in $M_4$.}. The average length of utterances is 9.32 in the whole PELD, which also conforms to the length of short sentences in daily conversation. 

Similar to existing emotional conversation datasets \cite{li2017dailydialog, busso2008iemocap}, PELD also suffers the emotion imbalance issue. Utterances labeled as \textit{Neutral} are the majority (44.6\%) , while \textit{Fear} and \textit{Disgust} only take a small portion (6.9\% and 1.9\%) . Though it reflects the actual emotion distribution in daily conversation, it also brings challenges to machine learning models to identify and generate emotions. We tried several automatic methods for data augmentation like synonym substitution, back-translation, or the Easy Data Augmentation (EDA) proposed in \cite{wei2019eda}. But most of the synthetic samples are either odd or the same as the original samples. The reason might be there are limited options for short sentences as utterances in conversation to replace synonyms, add or delete words.

The mood state distribution in PELD is the sum of emotions within mood states. All the triples are equally distributed among the six main roles, ranging from 977 (Phoebe) to 1159 (Rachel). So, there is no explicit dominant personality in our dataset, making the model more robust when trained on it.

\begin{table}[h]
    \centering
    \caption{Basic Statistics in PELD.}
    \linespread{1.2}
    \small
    \begin{tabular}{l|cccc} 
        \toprule
        \textbf{Basic Statistics} &\textbf{Train} & \textbf{Valid} & \textbf{Test} & \textbf{Total}\\	\hline
		 \#Triple & 5286 & 588 & 653 & 6527 \\
		 \#Unique Uttr. & 9273 & 1529 & 1679 & 10468\\
		 Avg. Uttr. Length & 9.26 & 9.33 & 8.95 & 9.32\\
		 \hline
		\textbf{\#Emotions} & \textbf{Train} & \textbf{Valid} & \textbf{Test} & \textbf{Total}\\
		\hline
		 Anger & 1857 & 238 & 247 & 2342\\
		 Disgust & 316 & 30 & 30 & 376\\
		 Fear & 1100 & 118 & 132 & 1350\\
		 Joy & 2883 & 321 & 345 & 3549 \\
		 Neutral & 7066 & 782 & 880 & 8728\\
		 Sadness & 1086 & 120 & 141 & 1347\\
		 Surprise & 1550 & 155 & 184 & 1889\\
		\hline
		\textbf{\#Mood States} & \textbf{Train} & \textbf{Valid} & \textbf{Test} & \textbf{Total}\\
		\hline
		 Neutral & 7066	 & 782 & 880 & 8728\\
		 $M_1$ & 4433 & 476 & 529 & 5438\\
		 $M_2$ & 3273 & 386 & 409 & 4068\\
		 $M_3$ & 1086 & 120 & 141 & 1347\\
		 $M_4$  & - & - & - & - \\
		\hline
		\textbf{\#Triples of Main Roles} & \textbf{Train} & \textbf{Valid} & \textbf{Test} & \textbf{Total}\\
		\hline
		 Chandler & 864 &  107 & 117 & 1088 \\
		 Joey & 929 & 96 & 100 & 1125 \\
		 Monica & 847 & 95 & 111 & 1053 \\
		 Phoebe & 789 & 90 & 98 & 977 \\
		 Rachel & 934 & 97 & 128 & 1159 \\
		 Ross & 923 &  103 & 99 & 1123 \\
		\bottomrule
    \end{tabular}
    \label{PELD_dataset}
\end{table}

\subsection{Mood Transitions in PELD}

After constructing PELD, we further explore the dataset in the aspect of mood transitions. As the triples in PELD are constructed for analyzing the transitions between $M_i$ in $U_1$ and the $M_r$ in $U_3$, we show both the mood states and emotion distributions in the $U_1$ and $U_3$ in Table \ref{Emo_Utter}, respectively. 

We can see that for both mood states and emotions, the distributions in $U_1$ and $U_3$ are similar, which means the transition of mood states and emotions are equitable in PELD triples, \textit{i.e.}, no explicit mood state or emotion transition dominates during the dialogues. It also conforms to the nature of daily conversations. Moreover, the proportions of all mood states and emotions are also similar to the overall statistics of PELD, which suggests that the mood states and emotions in PELD are also average distributed in the triples.

\begin{figure*}[t]
\centering
\includegraphics[trim={1cm 1cm 0 0},scale=0.36]{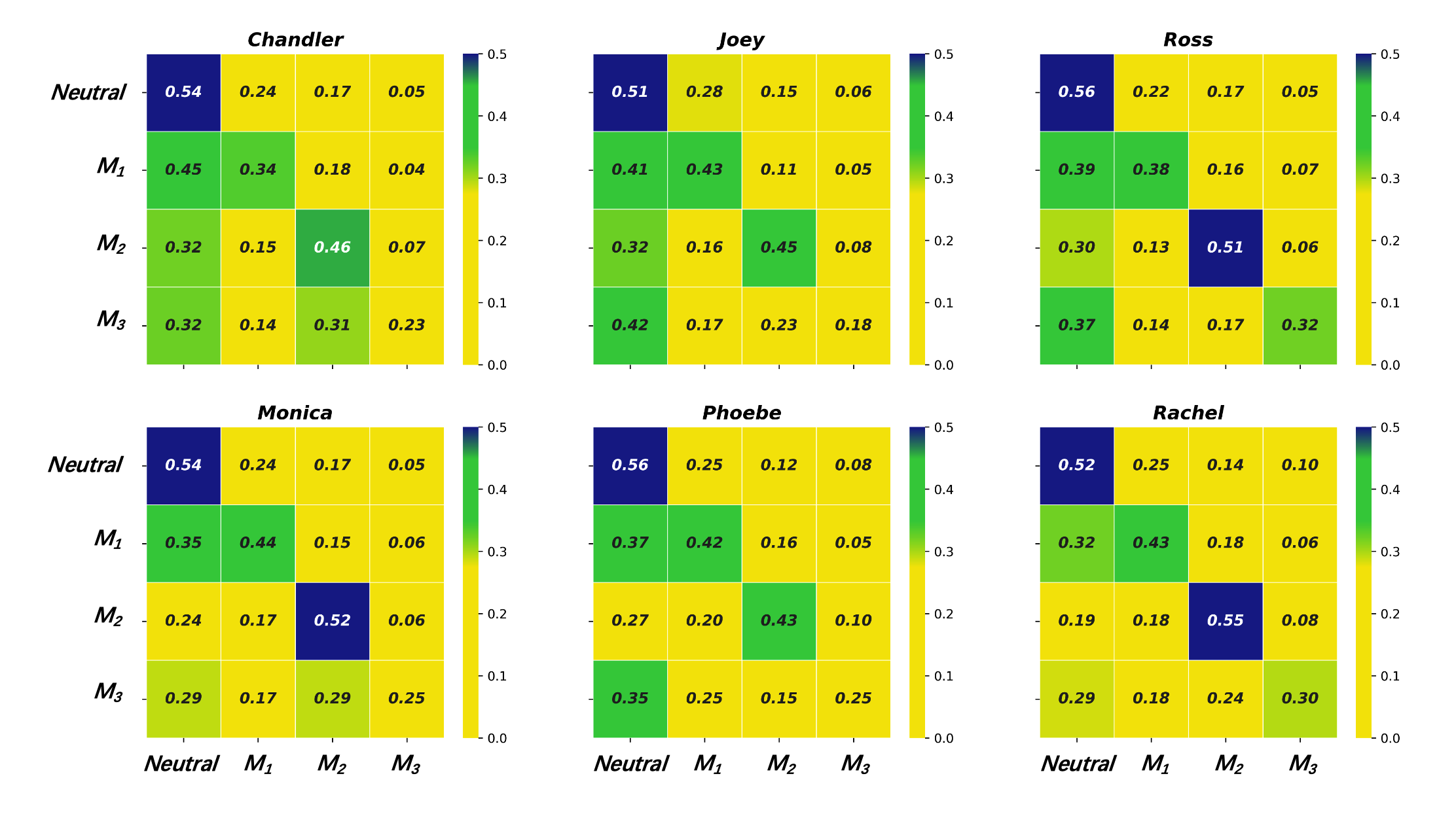}
\caption{Mood transition matrixes of the six main roles in PELD. Each row in a matrix shows the ratios of the current mood state $M_i$ is transferred to the next mood state $M_r$.}
\label{Mood_transition}
\end{figure*}

Since mood transitions are affected by the personality traits as discussed above, we exhibit the mood transition patterns for different roles with different personality traits in Figure \ref{Mood_transition}. In general, among the six transition matrixes, most of the first columns are in deeper colors, which indicates most transitions occur from other mood states to \textit{Neutral} as it is the majority in PELD. Besides, blocks with deeper color are also more likely to occur around or in the diagonals of the transition matrixes; it suggests the preceding mood states tend to transition to the same or similar ones.

\begin{table}[hb]
    \centering
    \caption{Mood States and Emotions distributions in PELD Triples}
    \linespread{1.2}
    \small
    \setlength\tabcolsep{3pt}
    \begin{tabular}{|c|c|c|c|c|c|c|c|} 
        \hline
        {Tri.Moods} &  {Neutral} & \multicolumn{2}{c|}{$ \rm M_1$}&\multicolumn{3}{c|}{$\rm M_2$}&\multicolumn{1}{c|}{$\rm M_3$} \\
         \hline
         \textbf{ $M_i$ } & 2922 & \multicolumn{2}{c|}{1842}&\multicolumn{3}{c|}{1325}&\multicolumn{1}{c|}{438} \\
         \textbf{ $M_r$ } & 2778 & \multicolumn{2}{c|}{1763}&\multicolumn{3}{c|}{1492}&\multicolumn{1}{c|}{494}\\
         \hline
         {Tri.Emos} &  {Neutral} & {Joy} & {Surprise} & {Anger}  & {Fear} & {Disgust} & {Sadness}\\
        \hline
         $E_i$ & 2922 & 1245 & 597 & 752 & 458 & 115 & 438 \\
         $E_r$ & 2778 & 1129 & 634 & 859 & 489 & 144 & 494 \\
        \hline
    \end{tabular}
    \label{Emo_Utter}
\end{table}

As for individual differences, nearly 60\% of $M_3$ in Joey and Phoebe transmit to $Neutral$ and $M_1$, which shows they can better handle the negative mood states. This also correlates to the lower Extraversion compared to others in Table \ref{personality_}. Specifically, over 80\% of $M_1$ in Joey remains to be $Neutral$ and $M_1$, which especially shows Joey is an optimistic person. As for the ratio of unpleasure mood states $M_2$ and $M_3$ change to pleasure mood state $M_1$, Phoebe achieves the highest.



\begin{figure}[t]
\centering
\includegraphics[trim={0 1cm 0 0}, scale=0.3]{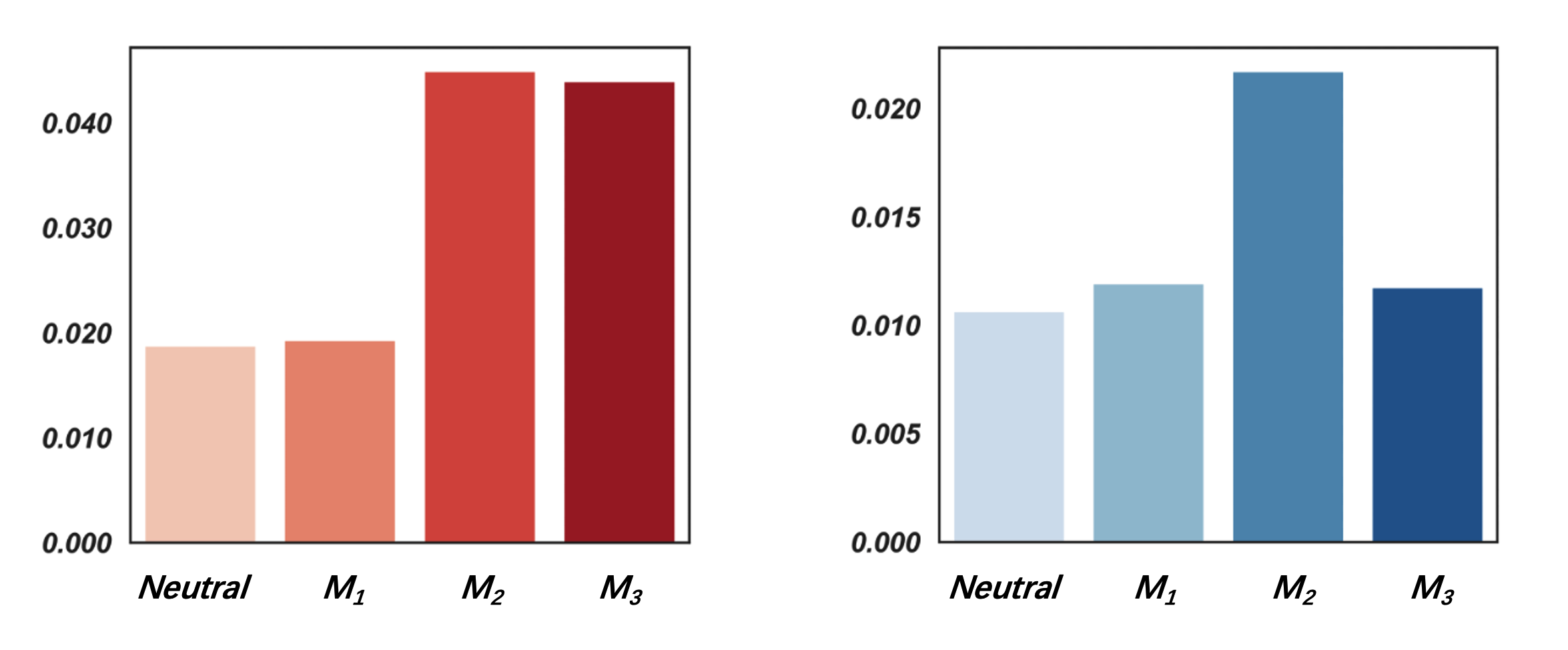}
\caption{The standard deviations of the infinity norm (red) and the L2-norm (blue) of each row in mood transition matrixes of the six main roles in PELD. In both figures, a higher bar indicates a greater disparity among the roles on the mood transition pattern in the corresponding row.}
\label{mood_transition_all}
\end{figure}

Moreover, to highlight the individual differences in mood transitions among the six main roles in detail, we also show the standard deviations (Std) of each row in the transition matrixes of the six main roles, as shown in Figure \ref{mood_transition_all}. The red bar chart shows the Std of the infinite norms of rows in the mood transition matrix, which indicates the diversity of the most probable mood states from the same mood state in transitions of different roles. While the blue bar chart shows the Std of the L2-norms, which generally describes the difference in how different roles transfer from one mood state to others. Detailed calculations of the standard deviations are shown below:

\begin{equation}
	\begin{aligned}
	Std(||\mathcal{M}^i||_2) &=  \sqrt{\frac{1}{m}\sum_{j=1}^{m}({\sum_{k=1}^{n}(|\mathcal{R}_{jk}|^2) }^{\frac{1}{2}} - \mu_{2}^i)^2}\\\
	Std(||\mathcal{M}^i||_{inf}) &= \sqrt{\frac{1}{m}\sum_{j=1}^{m}(\max||\mathcal{R}_{j}|| - \mu_{inf}^i)^2} \\\
	\end{aligned}
	\label{std}
\end{equation}

\noindent
where $||\mathcal{M}^i||_2$ is the L2-norm of the \textit{i}-th mood state among the six roles, while $||\mathcal{M}^i||_{inf}$ is the corresponding infinit-norm. $m$ is the number of the roles, $n$ is the number of columns in each mood transition matrix. $\mathcal{R}$ is the \textit{i}-th row in the mood transition matrix, indicating the result mood states of the \textit{i}-th mood state's transition. $\mu_{2}^i$ and $\mu_{inf}^i$ are the mean of $||\mathcal{M}^i||_2$ and $||\mathcal{M}^i||_{inf}$, respectively.

Both charts show similar patterns of mood transitions. Unpleasure moods ($M_2$ and $M_3$) vary the most in different roles, while people are more common when process \textit{Neutral} and pleasure mood $M_1$ in conversation. Besides, Unpleasure mood states ($M_2$ and $M_3$) are relatively higher than pleasure mood states and \textit{Neutral} on average, which means the individual difference is larger. So, we can infer that personality traits influence more in mood transitions from negative emotions.

\subsection{Personality-aware Mood State Transitions}
Although we have done an extensive analysis of the mood transitions of different characters, these conclusions may have certain limitations because their personality traits are specially designed by the scriptwriters for the TV series.

\begin{figure}[t]
\centering
\includegraphics[scale=0.5]{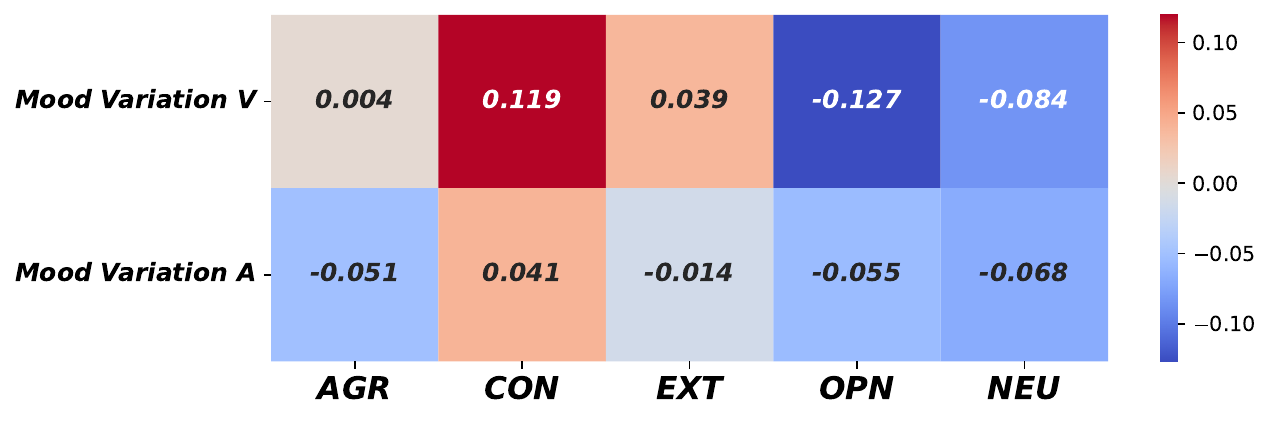}
\caption{Spearman correlations among mood state variations (from $U_1$ to $U_3$ in PELD's triples) and personality traits. The mood variations are in \textbf{V}alence and \textbf{A}rousal dimensions.}
\label{Mood_personality}
\end{figure}

In order to explore more of the influence of personality on mood transitions, we investigate the personality-aware mood state transitions. Specifically, we calculate the Spearman correlations between mood state transitions and different personality traits in PELD, as shown in Figure \ref{Mood_personality}.

Figure \ref{Mood_personality} shows the mood state variation from $U_1$ to $U_3$ in the triples in PELD, where we illustrate the variations in separate V-A dimensions\footnote{the mood states are only in V-A spaces as explained in Section 3}. We can see that Conscientiousness (C) has a higher positive correlation with mood state variation in the V-dimension. It suggests that the mood states of speakers with higher Conscientiousness tend to be more positive during conversations. In other words, if these speakers initially experience \textit{Fear}, \textit{Anger}, \textit{Sadness}, or \textit{Disgust}, they are more likely to transition toward \textit{Surprise} or \textit{Joy} as the conversation progresses. On the contrary, speakers with higher Openness (O) have a higher negative mood state variation in the V-dimension, which means these speakers are more likely to be first positive and then relatively negative in the conversations. In terms of emotion, these speakers are more likely to be first in \textit{Surprise} or \textit{Joy}, then become \textit{Fear}, \textit{Anger}, \textit{Sadness}, or \textit{Disgust}.

It is worth noting that the current conclusions we draw from analyzing mood state transition patterns based on the Big Five in PELD still have certain limitations. Personality expression, especially in conversational contexts, is affected by various cultural backgrounds, social factors, and even different conversation situations. Therefore, any analysis results from a single conversation dataset will have specific limitations.

However, our intent is not to derive a universal conclusion about the influence of personality on mood transitions in conversations from a social science perspective. The methods we present, such as calculating mood transition matrices, comparing the norms of these matrices, and examining the correlation between changes in different affective dimensions (V, A) and the intensity of different personality traits, can be used to analyze various conversation datasets with Big Five personality and affective annotations.


\section{Experiment}
\label{section:Experiment}
\subsection{Evaluation Task}

To validate the effectiveness of our proposed method, we conducted the Emotion Generation task on PELD. Emotion Generation requires the model to generate the appropriate response emotion in the upcoming utterance based on the preceding dialog context in a dyadic conversation scenario. The emotions here are discrete labels of basic emotions (\textit{i.e.}, \textit{Anger}, \textit{Disgust}, \textit{Fear}, \textit{Joy}, \textit{Neutral}, \textit{Sadness}, and \textit{Surprise}).

We evaluate the performance by F-scores of single emotions. Besides, the overall performance is also measured from two aspects: the macro averaged F-score (\textbf{m-avg}), and the weighted averaged F-score (\textbf{w-avg}) of all seven emotions. A higher m-avg indicates the model performs relatively better in predicting each category, while a higher w-avg means the model predicts mood states or emotions with larger proportions in the dataset better. Our evaluation is under the assumption that emotions expressed in the upcoming utterance are the emotions generated by the speakers. 

\subsection{Ablation Study Setting}
Although there are existing studies working on emotion recognition in conversation \cite{lin2022modeling,song2022supervised}, there is no existing model to solve the Emotion Generation task issued in our work. Therefore, we conduct an ablation study to evaluate the effectiveness of different modules of our model design and the ways we utilize the personality and model the mood transition. The ablation study compares the performances of the following baseline models: \\

\noindent
\textbf{BERT-base:} BERT-base \cite{devlin2018bert} is a famous pre-trained language model with 12 transformer layers, over 110 million parameters, and 12 attention heads. It is capable of capturing bidirectional context for performing a wide range of downstream NLU tasks. Bert-base is pre-trained with a large corpus of text consisting of around 3.3 billion words and 2,500 million sentences. This vast corpus allowed BERT to learn a wide range of linguistic patterns and structures, making it a highly effective pre-trained model for natural language processing tasks. We use the pre-trained BERT-base model, corresponding to the $E_n$ in our model, to encode the preceding dialog context to obtain the semantic representation as input, then directly predict the emotions for response through a classification head. \\

\noindent
\textbf{BERT-Mood:} based on the vanilla BERT-base model, we add a mood state classification as the auxiliary task to enhance Emotion Generation. In BERT-Mood, we first use the BERT-base model to encode the preceding dialog context to obtain the semantic representation. Then, the semantic representation is fed into two classification heads: Emotion Generation and mood state classification. The sum of the losses from these two classification tasks will be back-propagated to the BERT-base model and update its parameters. This baseline model is to verify if the mood state information helps the Emotion Generation task.\\

\noindent
\textbf{BERT-MT:} BERT-MT (Mood Transition) integrates the mood transition module (as described in Section 5.2) into BERT-base. In BERT-MT, the transitioned mood state is concatenated with the dialog context to help generate $E_r$, where we use both the regression loss of the mood state generation and the classification loss of Emotion Generation for supervision. In this baseline model, the personality is ablated to its effectiveness. To eliminate the influence of the model scale, we keep the number of parameters the same as our model and randomly initialize the personality-related parameters.\\

\noindent
\textbf{BERT-P:} In BERT-P, we concatenate the preceding context representation from BERT-base and the specified personality trait vector together. Then, the concatenated representation is fed into the classification head for Emotion Generation. This baseline model is to evaluate whether merely using the personality trait vector can enhance Emotion Generation.\\

\subsection{Implementation Details}
To facilitate the reproduction of our model, we add the implementation details as follows. The dialog flows in PELD are fed into the models in batches with a size of 16. In encoding the preceding dialog context, we pad all the utterances with [PAD] to the MAX\_LEN of 128. We adopt the pre-trained BERT-base model from huggingface\footnote{huggingface.com}.

We set the warm-up as 0.05 of the total training times. Besides, we use the Adam \cite{kingma2014adam} as the optimization algorithm in training. The learning rate for all the models is set as \textit{1e-5}. All the models for testing are selected referring to the best performance on the Validation sets in 50 epochs of training. 

To ensure the reliability of results, we run each experiment 10 times with different random seeds and record the average performances and the stardard deviations, where the random seeds are used to split the datasets and initialize the model parameters. Lastly, we released our code and data at github\footnote{https://github.com/preke/PELD}.

\section{Results and Analysis}
\label{section:Result}
In this section, we report and analyze the experimental results of Emotion Generation in our ablation study by answering several research questions:

\begin{enumerate}[label=\bfseries RQ\arabic*:,leftmargin=.5in]
    \item Does our model outperform baseline methods in Emotion Generation? 
    \item What is the correlation between the mood transition process and Emotion Generation?
    \item Are certain personalities easier to predict mood transition and Emotion Generation?
    \item Are there any cases to show how our model generates the emotions for the response?
\end{enumerate}


\subsection{Does our model improve baseline methods with the mood transition and personality in Emotion Generation?}

We report the performance of Emotion Generation on the test set of the PELD in Table \ref{Emotion_result}. To show the significance of the outperformance of our method, we conduct the Welch's t-test \cite{welch1947generalization} between our model and all the baseline models. The Welch's t-test, also known as Welch's unequal variances t-test, is a statistical test used to compare the means of two independent groups when the assumption of equal variances is violated. It is a modification of the traditional Student's t-test that allows for unequal variances between the groups being compared. We employed Welch's t-test to assess the statistical significance of our model's superior performance over other baseline models. Since these results were derived from different models, we cannot guarantee that their variances are equal or similar. Therefore, we opted for Welch's t-test instead of the standard t-test.


\begin{table*}[t]
    \centering
    \caption{Result comparison among baseline methods and our model in Emotion Generation. The first lines in baseline results are the F-scores of Emotion Generation (averaged over ten runs), while the second lines are the P-values $p$ by calculating Welch's t-test between each baseline and our method. The green results indicate that our model outperforms the corresponding baselines with $p \leq 0.05$, where we consider the gain in performance to be statistically significant.}
    \linespread{1.2}
    \small
    \begin{tabular}{c|ccccccccc} 
        \toprule
        \textbf{Methods} & \textbf{Anger} & \textbf{Disgust} & \textbf{Fear} & \textbf{Joy} & \textbf{Neutral} & \textbf{Sadness} & \textbf{Surprise} & \textbf{m-avg} & \textbf{w-avg} \\
            \hline
            \multirow{2}{*}{BERT}
             & \textcolor[RGB]{10, 160, 46}{0.318} & \textcolor[RGB]{10, 160, 46}{0.012} & 0.226 & \textcolor[RGB]{10, 160, 46}{0.278} & \textcolor[RGB]{10, 160, 46}{0.513} & \textcolor[RGB]{10, 160, 46}{0.212} & \textcolor[RGB]{10, 160, 46}{0.109}  & \textcolor[RGB]{10, 160, 46}{0.242} & \textcolor[RGB]{10, 160, 46}{0.375} \\
             &\textcolor[RGB]{10, 160, 46}{0.05} &\textcolor[RGB]{10, 160, 46}{0.02} & 0.29 & \textcolor[RGB]{10, 160, 46}{0.05} & \textcolor[RGB]{10, 160, 46}{0.03} & \textcolor[RGB]{10, 160, 46}{0.03}  & \textcolor[RGB]{10, 160, 46}{0.03} & \textcolor[RGB]{10, 160, 46}{0.03}  & \textcolor[RGB]{10, 160, 46}{0.03}  \\
            \cline{1-10}
            \multirow{2}{*}{BERT-Mood}
             & \textcolor[RGB]{10, 160, 46}{0.252} & \textcolor[RGB]{10, 160, 46}{0.113} & 0.227 & \textcolor[RGB]{10, 160, 46}{0.248} & \textcolor[RGB]{10, 160, 46}{0.468} & 0.288 & \textcolor[RGB]{10, 160, 46}{0.107}  & \textcolor[RGB]{10, 160, 46}{0.242} & \textcolor[RGB]{10, 160, 46}{0.344} \\
             &\textcolor[RGB]{10, 160, 46}{0.01} & \textcolor[RGB]{10, 160, 46}{0.00} & 0.36 &\textcolor[RGB]{10, 160, 46}{0.03} &\textcolor[RGB]{10, 160, 46}{0.00} & 0.05  &\textcolor[RGB]{10, 160, 46}{0.00} & \textcolor[RGB]{10, 160, 46}{0.01} &\textcolor[RGB]{10, 160, 46}{0.00} \\
             \cline{1-10}
            \multirow{2}{*}{BERT-P} 
            & \textcolor[RGB]{10, 160, 46}{0.267} & \textcolor[RGB]{10, 160, 46}{0.096} & \textcolor[RGB]{10, 160, 46}{0.159} & 0.320 & \textcolor[RGB]{10, 160, 46}{0.494} & \textbf{0.299} & 0.119 & \textcolor[RGB]{10, 160, 46}{0.254} & \textcolor[RGB]{10, 160, 46}{0.349} \\
            &\textcolor[RGB]{10, 160, 46}{0.05} & \textcolor[RGB]{10, 160, 46}{0.04} & \textcolor[RGB]{10, 160, 46}{0.05} & 0.01 &\textcolor[RGB]{10, 160, 46}{0.05} & 0.03       & 0.01 &\textcolor[RGB]{10, 160, 46}{0.05} & \textcolor[RGB]{10, 160, 46}{0.03}       \\
            \cline{1-10}
            \multirow{2}{*}{BERT-MT} 
             & \textcolor[RGB]{10, 160, 46}{0.271} & \textcolor[RGB]{10, 160, 46}{0.099} & 0.173 & \textbf{0.334} & \textcolor[RGB]{10, 160, 46}{0.507} & \textcolor[RGB]{10, 160, 46}{0.239} & \textbf{0.127} & 0.247 & \textcolor[RGB]{10, 160, 46}{0.368} \\ 
             & \textcolor[RGB]{10, 160, 46}{0.05} & \textcolor[RGB]{10, 160, 46}{0.02} &0.40 & 0.03 & \textcolor[RGB]{10, 160, 46}{0.04} &\textcolor[RGB]{10, 160, 46}{0.03}  & 0.02 & 0.39 & \textcolor[RGB]{10, 160, 46}{0.04} \\
             \cline{1-10}
            Our Model & \textbf{0.323} & \textbf{0.167} & \textbf{0.229} & 0.291 & \textbf{0.545} & 0.254 & 0.114  & \textbf{0.269} & \textbf{0.392} \\
        \bottomrule
    \end{tabular}
    \label{Emotion_result}
\end{table*}

As we can see in Table \ref{Emotion_result}, the comprehensive performance of all models is moderately low, which indicates the task's difficulty. The reason is that Emotion Generation is a seven-classes generation task without knowing the response content. Besides, it also suffers from the imbalance issue, as shown in the data distribution in Table \ref{Emo_Utter}. When we look at the average F-scores, w-avgs of all models are higher than the m-avgs. It also verifies that generating the majority emotion (\textit{i.e., Neutral}) is easier than other emotions. Our model statistical-significantly outperforms all other baseline models in w-avg, \textit{Anger}, \textit{Disgust}, and \textit{Neutral}. Besides, it also surpasses multiple baseline models in m-avg, \textit{Surprise}, \textit{Joy}, and \textit{Sadness}. The results verify that our model design efficiently utilizes the mood transition process and the personality to improve Emotion Generation.

%

The lowest performance occurs when we only use the BERT-base model for the Emotion Generation task. The m-avg is \textbf{0.242} and the F-score of \textit{Disgust} is only \textbf{0.012}. However, when we integrate the mood state transition and the personality trait into the BERT model, the performances are immediately improved, especially in generating \textit{Disgust}. It suggests that our method improves the base model by raising the performance of generating minority emotions.

We first compare the performances of BERT, BERT-Mood, and BERT-MT to illustrate how different utilizations of mood states influence Emotion Generation. By comparing BERT and BERT-Mood, we can see that integrating the mood transition prediction task decreases the F-scores of the majority of emotions like \textit{Anger, Joy} and \textit{Neutral} but increase the rest of minorities. Consequently, although the m-avg remains the same, the w-avg goes down. So, predicting the mood state helps generate minority emotions.

Then we look at BERT-MT compared with BERT-Mood, F-scores of  \textit{Fear}, \textit{Sadness}, and \textit{Fear} decrease, while the F-scores of all other emotions grow up. Besides, both the m-avg and the w-avg improve than BERT-Mood. So, integrating the mood transition process in Emotion Generation fixed the performance issue of the majority of emotions in BERT-Mood.

Then, we focus on the personality. Comparing BERT-P and BERT, we can see that: although integrating the personality trait can slightly improve the m-avg, some majority emotions, e.g., \textit{ Neutral, Anger} are lower so that the w-avg also decreases. However, when utilizing the personality as the mood transition weight in our model, we can see the performance is much better. So, in the Emotion Generation task, directly concatenating the personality trait causes unstable improvement, but modeled as the mood transition weight, the personality helps achieve robust and significant enhancement.

\subsection{What is the correlation between the mood transition process and Emotion Generation?}

\begin{figure}[t]
\centering
\includegraphics[scale=0.58]{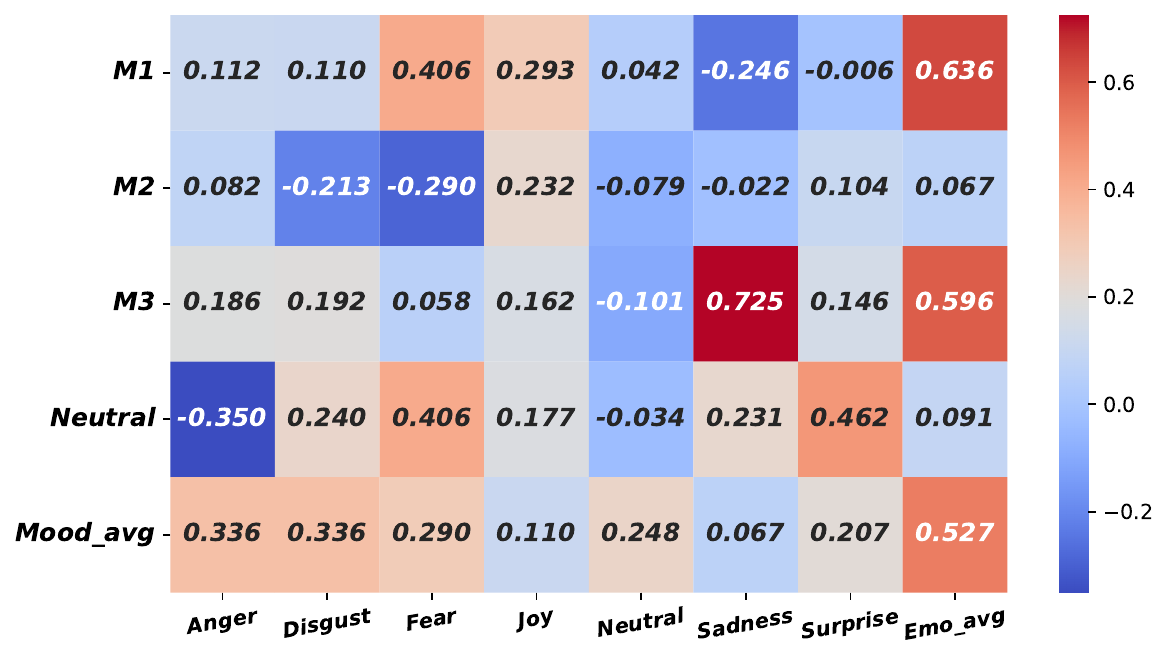}
\caption{Spearman correlations among F-scores of mood transition prediction and Emotion Generation. }
\label{corr}
\end{figure}

To elucidate the correlation between the mood transition process and Emotion Generation, we present the impact of distinct mood transition processes on the generation of diverse emotions. Specifically, we conduct experiments of Emotion Generation with 10 random seeds and calculate the Spearman correlation coefficient between the results of mood transition (the F-scores of accurately predicting each mood in the response) and the F-scores of different emotions. The correlation is illustrated in Figure \ref{corr}. In particular, the last row in Table \ref{corr} indicates the Spearman correlation coefficient between the average F-scores of all the mood transition predictions and the F scores of each emotion generation. Similarly, the last column in Table \ref{corr} is the Spearman correlation coefficient between the average F-scores of all emotion generations and the F scores of each mood transition prediction.

In general, Emotion Generation is partially conditioned on the results of mood transition; accurate prediction of mood transition means that the information provided to Emotion Generation is correct. The last row of Figure 8 shows that the overall predictions for mood transitions are positively correlated with the generation of all emotions. Besides, the last column shows that the transition predictions of all single mood states are also positively correlated with Emotion Generation.

The last column shows that the F-score of predicting \textit{M1} has the highest correlation with the average result of Emotion Generation. We infer the reason is that emotions in \textit{M1} (\textit{Joy} and \textit{Surprise}) take a large proportion of all emotions in PELD. Accurately predicting \textit{M1} helps to correctly generate these emotions. Moreover, the high correlation between \textit{M3} and \textit{Emo\_avg} shows that \textit{M3} only corresponds to the minority emotion \textit{Sadness}, and correctly predicting \textit{M3} will help eliminate interferences in Emotion Generation.

We further look at the correlations among specific mood states and emotions. The accurate prediction of \textit{M3} is highly correlated with the generation of \textit{Sadness}, and it is the highest correlation value among all. This is also due to  \textit{M3} only corresponding to one emotion \textit{Sadness}, as mentioned before. However, we also noticed that the results of some mood transitions are negatively correlated with Emotion Generation, such as \textit{Neutral} mood states and \textit{Anger}, or \textit{M2} and \textit{Fear}. It indicates that when these mood transitions are accurately predicted, some Emotion Generation will be decreased. We infer the reason behind this might be that these two different losses $\mathcal{L}_{mood}$ and $\mathcal{L}_{emo}$ in our model inevitably conflict with each other. However, because the introduction of mood transition has improved the Emotion Generation as a whole, as mentioned above, we retain the current design. 

\subsection{Do certain personalities easier to predict mood transition and Emotion Generation}

\begin{figure}[t]
\centering
\includegraphics[scale=0.65]{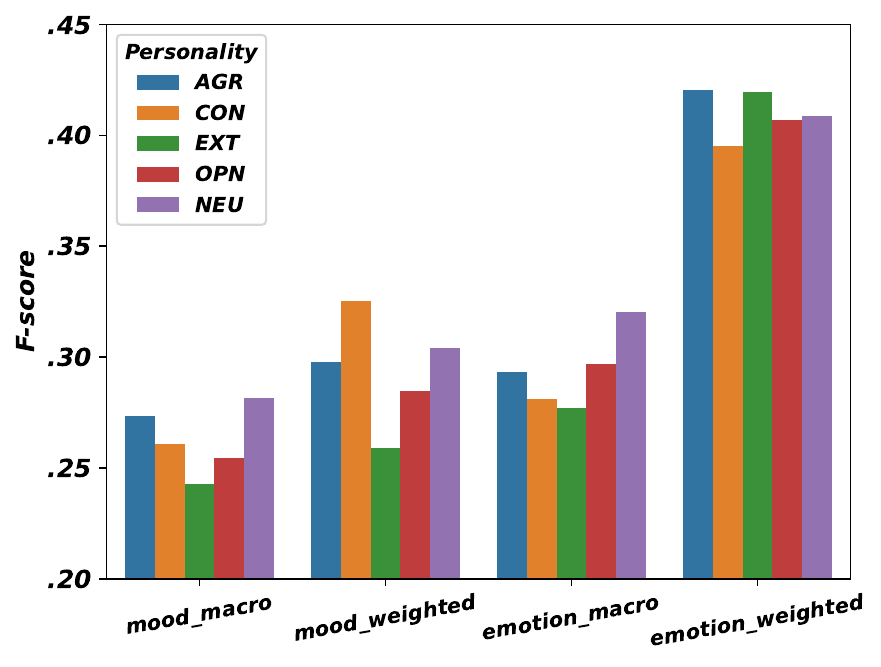}
\caption{This bar plot illustrates the F-scores of predicting Mood Transition (mood\_macro and mood\_weighted) and Emotion Generation (emo\_macro and emo\_weighted) for speakers with different personalities using our method.}
\label{corr_2}
\end{figure}

We analyze the results of our method (averaged from 10 random seeds) on the test set based on different personalities. The results are shown in Figure \ref{corr_2}.

Specifically, for the triples in the test set, we manually labeled the speakers with binary big-five personality categories according to the original annotations in FriendsPersona. It means if the triples occur in FriendsPersona, we directly adopt the annotations; otherwise, we manually label the personality categories of the speakers.

The results in Figure \ref{corr_2} show that the mood transition results of different personalities (both mood\_macro and mood\_weighted) vary a lot. The mood states of speakers with NEU are easier to predict, while speakers with EXT are more difficult to be captured, relatively. We infer the reason might be: NEU speakers are more sensitive in conversation. Their mood states will be easier to be influenced by the dialog content. Therefore, the dialog content as the mood transition variables in our method provides more valuable information. While EXT speakers tend to be outgoing and talkative in most dialog contexts, their subtle mood states transitions are more difficult to capture in our method.

As for the Emotion Generation, emotion\_macro of different personalities also vary, but emotion\_weighted of different personalities are more similar to each other. The reason is that correctly predicting the majority emotion \textit{Neutral} is counted more in calculating emotion\_weighted compared with emotion\_macro. However, \textit{Neutral} is expressed the most among all speakers regardless of their personalities. Besides, the patterns of speakers of EXT and NEU remain similar in emotion\_macro due to their characteristics.

\subsection{Is there any cases to show how our model generate the emotions for the response?}

We conduct a case study and analyze the result samples of our method on the test set to show how our method works in real conversation scenarios. As shown in Table \ref{case_study}, the Utterance 1 and the Ground Truth Response are from the same speaker with a known personality.

First, we show two representative samples where our model correctly generates the emotion for the response. The first sample occurs when the two speakers are talking about the fetal movement, both speakers are in the \textit{Joy} emotion in the context, so our model generates the \textit{Joy} for the response and is verified in the ground truth. In the second sample, Rachel is sad in utterance 1; after being comforted and helped by Michael in utterance 2, she becomes happy in the response. This sample verifies that our model is able to generate the \textit{Joy} emotion with the personality of Rachel. With relatively higher Extroversion, Rachel's mood state is easier to be influenced by others and express emotion with higher Arousability.

Moreover, we also show an example where our model makes mistakes. In the third case, our model wrongly generates the \textit{Angry} because the two speakers argue angrily in the context. However, the content of Rachel's response clearly demonstrates her willingness to attend the lecture. This case also illustrates the limitation of our method: without knowing and modeling enough background knowledge of the speakers, it is difficult to generate appropriate emotion in conversations.

\begin{table}[t]
    \centering
    \caption{Emotion generation illustration on dialog samples in PELD.}
    \linespread{1.2}
    \small
    \begin{tabularx}{\textwidth}{|X|X|p{40pt}|X|} 
            \hline
        \makecell*[c]{\textbf{Utterance 1}}  & \makecell*[c]{\textbf{Utterance 2}} & \textbf{Generated Emotion} & \makecell*[c]{\textbf{Ground Truth Response}}  \\
            \hline
            \textbf{Joey:} Oh, yeah. he's got that great baby smell. get a whiff of his head. (\textbf{\textit{Joy}}) & \textbf{Caroline:} I think my uterus just skipped a beat! ( \textbf{\textit{Joy}}) & \makecell*[c]{\textbf{\textit{Joy}} } &\textbf{Joey:} What'd I tell you? What'd i tell you? ( \textbf{\textit{Joy}})\\
            \hline
            \textbf{Rachel:} Oh, look at me, look at me. Oh, I'm on a date with a really great guy, all I can think about is Ross and his cat and his... Julie. I just want to get over him. God, I just, why can't I do that? ( \textbf{\textit{Sadness}}) & \textbf{Machael:} Oy. Look, I've been through a divorce, trust me, you're gonna be fine. You just can't see it now because you haven't had any closure. (\textbf{\textit{Sadness}}) & \makecell*[c]{\textbf{\textit{Joy}} } & \textbf{Rachel:} Yeah! Closure. That's what it is, that's what I need. God, you're brilliant! ( \textbf{\textit{Joy}} )\\
            \hline
            \textbf{Rachel:} Is that funny? Am I supposed to be laughing? (\textbf{\textit{Anger}}) &
            \textbf{Ross:} I don't know, you thought 'See you Saturday' was funny. Look honey, Mark is in fashion okay, I like having a friend that I can share this stuff with. You guys would never want to go to a lecture with me. (\textbf{\textit{Anger}}) & \makecell*[c]{\textbf{\textit{Anger}} } & \textbf{Rachel:} Pa-haa!! I would love to go with you. (\textbf{\textit{Joy}}) \\
            \hline
    \end{tabularx}
    \label{case_study}
\end{table}

\section{Conclusion and future work}
\label{section:Conclusion}
In this work, we raise a new task of personality-affected emotion generation and propose a new perspective to solve it through personality-affected mood transition. Besides, we construct a dialog script dataset PELD with emotion and personality labels to facilitate related research. We conduct extensive experiments on PELD to evaluate the effectiveness of our method. The results verify that integrating the personality and the mood transition regression significantly improves the performance in emotion generation, especially in minority emotions.
 
In future research, we intend to focus on two issues: (1) the personality effects on emotions in the multi-modality scenario, and (2) personality effects on response generation.
Facial expressions, voices, gestures, and environment information are also vital in emotional interaction, but they are not captured in purely text-based dialog systems. Besides, as seen from statistics in PELD, the most common emotion in the dialog scripts is still \textit{Neutral}. One possible reason is that other subtle affective information is not captured in the text. Therefore, our future works will continue to investigate the personality effects on emotions in the multi-modality scenario. Besides, the influence of personality on language usage is also studied in existing works \cite{tausczik2010psychological,cannava2018stuff}. To construct intelligent dialog systems with personality, it is also important to investigate how the given personality influences the semantic content in response generation \cite{huang2020challenges}.

\section{Acknowledgement}
This work is conducted at the Research Institute for Artificial Intelligence of Things (RIAIoT) at PolyU and supported by the GRF-RGC General Research Fund 2021/22 (No.15204921), and financial supported by Lingnan University (LU) (DB23A4) and Lam Woo Research Fund at LU (871236).


\bibliographystyle{ACM-Reference-Format}
\bibliography{reference}


\end{document}